\documentclass[10pt,twocolumn,letterpaper]{article}

\usepackage[pagenumbers]{cvpr} 

\definecolor{cvprblue}{rgb}{0.21,0.49,0.74}
\usepackage[pagebackref,breaklinks,colorlinks,allcolors=cvprblue]{hyperref}

\usepackage{multirow}
\usepackage{multicol}
\usepackage{makecell}

\def\paperID{11263} 
\def\confName{CVPR}
\def\confYear{2025}

\title{ReCon: Enhancing True Correspondence Discrimination through Relation Consistency for Robust Noisy Correspondence Learning}

\author{Quanxing Zha$^1$,~~Xin Liu$^{1,2}$\thanks{Corresponding author},~~~Shu-Juan Peng$^1$,~~Yiu-ming Cheung$^{2}$,~~Xing Xu$^3$,~~Nannan Wang$^4$\\
$^1$Huaqiao University, $^2$Hong Kong Baptist University\\
$^3$University of Electronic Science and Technology of
China, $^4$Xidian University\\
{quanxing.zha@gmail.com, xliu@hqu.edu.cn}
}

\begin{document}
\maketitle

\begin{abstract}
Can we accurately identify the true correspondences from multimodal datasets containing mismatched data pairs? Existing methods primarily emphasize the similarity matching between the representations of objects across modalities, potentially neglecting the crucial relation consistency within modalities that are particularly important for distinguishing the true and false correspondences. Such an omission often runs the risk of misidentifying negatives as positives, thus leading to unanticipated performance degradation. To address this problem, we propose a general \textbf{Re}lation \textbf{Con}sistency learning framework, namely \textbf{ReCon}, to accurately discriminate the true correspondences among the multimodal data and thus effectively mitigate the adverse impact caused by mismatches. Specifically, ReCon leverages a novel relation consistency learning to ensure the dual-alignment, respectively of, the cross-modal relation consistency between different modalities and the intra-modal relation consistency within modalities. Thanks to such dual constrains on relations, ReCon significantly enhances its effectiveness for true correspondence discrimination and therefore reliably filters out the mismatched pairs to mitigate the risks of wrong supervisions. Extensive experiments on three widely-used benchmark datasets, including Flickr30K, MS-COCO, and Conceptual Captions, are conducted to demonstrate the effectiveness and superiority of ReCon compared with other SOTAs. The code is available at: \href{https://github.com/qxzha/ReCon}{https://github.com/qxzha/ReCon}.


\end{abstract}


\section{Introduction}
Cross-modal retrieval is dedicated to understanding the semantic correspondences between multimedia data, aiming to recall the most relevant candidates for a given query \cite{scan,cmgm,dscmr,pecmr}. While existing approaches have achieved remarkable success by associating the heterogeneous data in a common latent space, they often neglect to provide an explicit consideration of semantically irrelevant data. Such mismatches, a.k.a., \textit{noisy correspondence} (NC) \cite{ncr}, would be inadvertently introduced due to the notoriously labor-intensive data collection and the unreliable non-expert annotations \cite{cc, scaling}, which inevitably impedes the semantic correspondences between modalities and consequently results in a decline of retrieval performance \cite{BiCro,crcl,MSCN}. 

\begin{figure}
    \centering
    \includegraphics[width=0.98\linewidth]{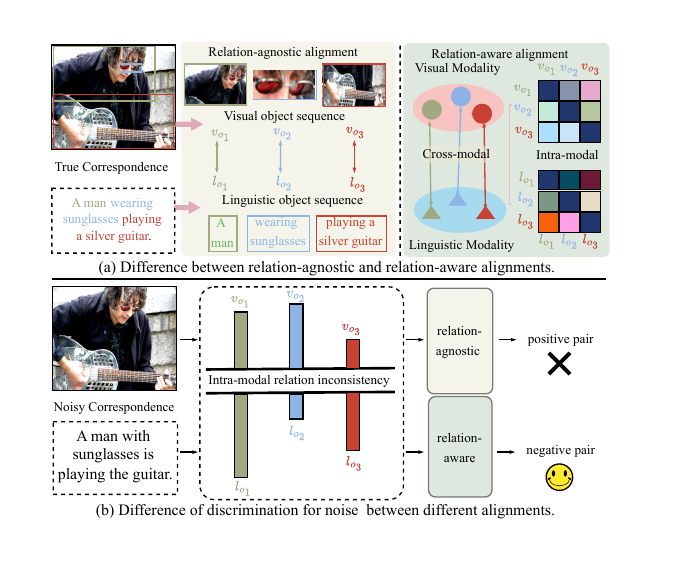}
    \caption{Illustration of relation discrepancy. The relation-aware alignment correctly identifies mismatched pair as negatives, while relation-agnostic alignment fails to detect such inconsistency.}
    \label{motivation}
\end{figure}

To tackle the NC problem, a core consensus is to enhance the discriminability for positives/matches and to mine local correspondences from negatives/mismatches. Several priors \cite{ncr,MSCN,BiCro} leverage the \textit{memory effect} \cite{coteaching}, wherein DNNs learn simple dominant patterns first, to identify matches in the early training stage. Subsequently, they estimate soft correspondence labels to describe the matching degree of mismatches, thus down-weighting their contributions and enforcing learning the local correspondences. In order to avoid the misleading caused by easily-determined noisy pairs, some attempts \cite{trip,cream,UGNCL} propose more refined data division strategies to filter out these mismatches. To further mitigate the wrong supervisions of mismatches, recent efforts \cite{l2rm,PC2} are presented to utilize pseudo counterparts for these mismatches to excavate informative correspondences. Furthermore, some works \cite{esc,gsc} achieve notable performance improvements by leveraging intrinsic properties observed within data, and methods \cite{decl,rcl,crcl} based on robust loss functions also effectively confront with the challenge of NCs. Nevertheless, they all neglect the relations within modalities, risking the misidentification of negatives as positives, particularly in cases of mismatched pairs that manifest high similarity scores, i.e., hard NCs. 


 As mentioned in IAIS work~\cite{iais}, the relation consistency often enhances the contextualized representation of image-text pairs. Inspired by this finding, we consider the relation discrepancy to mitigate the adverse impacts caused by mismatches among dataset.  As shown in Fig. \ref{motivation}(a), whether through relation-agnostic or relation-aware alignment, the true correspondence is expected to consistently assigned a high similarity score due to its perfect matching of both objects across modalities and relations within modalities. However, such matching is irreversibly compromised by the presence of untouchable noisy correspondence, thus narrows the distance between mismatches. Specifically, the unwanted misalignment erroneously reduces the distance of unassociated objects, inadvertently confusing retrieval models and thus undermining their discriminability for true correspondences. Besides, such misalignment also impairs the relations between objects within modalities, which significantly disrupts the contextual semantic consistency that is essential for true correspondences despite the nuance from objects. As shown in Fig. \ref{motivation}(b), the noisy pair with similar objects cannot be correctly identified by the relation-agnostic alignment due to its inability to recognize the discrepancies of relations within modalities. Such misidentification inevitably introduces false supervisions, which misleads the model towards further wrong optimization direction. In contrast, the relation-aware alignment accurately identifies it as a negative pair, benefiting from its dual consideration of both cross-modal and intra-modal relations.

\begin{figure*}
    \centering
    \includegraphics{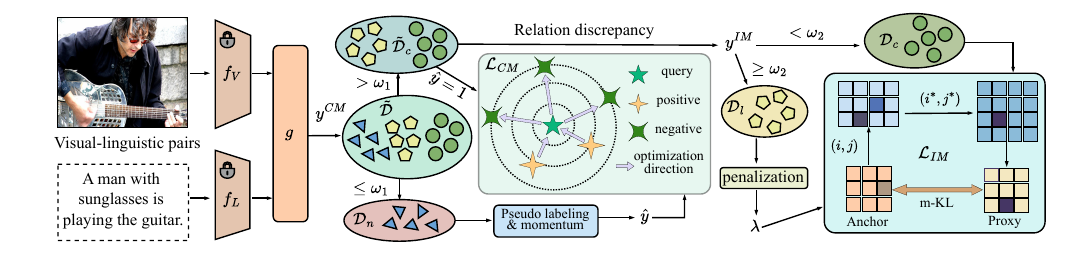}
    \caption{The schematic pipeline of the proposed ReCon learning framework.}
    \label{fig:network}
\end{figure*}

Motivated by the above observations, we propose a general \textbf{Re}lation \textbf{Con}sistency learning framework, namely \textbf{ReCon}, to effectively mitigate the adverse impact caused by NCs, as shown in Fig.~\ref{fig:network}. The main motivation of ReCon is to \textit{enhance the discriminability for true correspondences}. Specifically, an effective relation consistency alignment strategy is introduced to enable alignment not only between objects across modalities but relations within modalities. In details, the cross-modal relation consistency is presented to maximize the similarity scores of positive pairs while minimizing the negatives, ensuring that aligned objects have similar semantic representations. Meanwhile, the intra-modal relation consistency is employed to minimize the distance of relation matrices that describe the contextualized semantics of objects within modalities, which further enlarges the distinguish between positives and negatives. In practice, due to the lack of explicit annotations of objects, we propose to align the relation matrix extracted from one selected anchor modality with the proxy relation matrix extracted from another modality. Subsequently, such dual constraints of relations are employed to divide the noisy training data, wherein the divided partitions will be trained with corresponding strategies to achieve robust cross-modal retrieval, which significantly enhances the discriminability for true correspondences and effectively avoids the wrong supervisions of misidentified negatives. 


In summary, the main contributions of our work are as follows: (1) A general \textbf{Re}lation \textbf{Con}sistency learning framework, namely \textbf{ReCon}, is robustly proposed to identify the true correspondences and therefore mitigate the adverse impact caused by NCs within multimodal dataset. (2) An effective relation consistency alignment strategy is explicitly employed to jointly enforce the alignment of the cross-modal relation consistency and the intra-modal relation consistency. (3) A reliable true correspondence discrimination strategy is effectively presented to accurately partition the noisy data pairs, which therefore, seamlessly minimizes the risk caused by wrong supervisions and mitigates the misidentification of mismatches. (4) Extensive experiments highlight the advantages of the proposed ReCon in comparison with other SOTA methods and demonstrate its outstanding performances in challenging NCs scenario.



\section{Related Work}
\subsection{Cross-Modal Retrieval}
Cross-modal retrieval (CMR) aims to search the most relevant items across different modalities in response to query modality. The core of CMR is to minimize the semantic discrepancies by projecting different modalities into a common comparable space, wherein the matched items manifest higher similarity or closer feature distance and vice versa. Current efforts, from the perspective of similarity calculation, can be roughly classified into two categories: 1) Coarse-grained measurement \cite{vse++,vsrn,gpo,vtsr}, which represents an efficient solution with the key idea of associating the correspondence holistically among features extracted by distinct modality-specific encoders. 2) Fine-grained measurement \cite{scan,naaf,cmgm, sgraf, lsra, cross-graph,chan}, which focuses on assessing the semantic relationships at a more granular level to learn and reason latent alignments between fragments. Unfortunately, the promising performance of all these methods relies heavily on an implicit assumption that all training data pairs are correctly matched while neglecting the presence of NC. Such NC inevitably undermines the alignments and complicates the accurate measurement of similarity, ultimately leading to inferior performance.

\subsection{Noisy Correspondence Learning}
Noisy correspondence learning refers to noise-tolerant approaches well-designed to effectively mitigate the adverse impacts caused by mismatches among dataset. Unlike traditional category-level mistaken annotations, this instance-level semantic inconsistency, first recognized as a new paradigm of noisy labels in \cite{ncr}, significantly affects the performance of retrieval models. Thus, some prior attempts \cite{ncr, BiCro,MSCN} employ the small-loss criterion \cite{dividemix} to identify matched pairs from the corrupted datasets and subsequently rectify soft correspondence labels for those mismatches. Following this, several works \cite{cream,trip,UGNCL} introduce novel criteria to enable more fine-grained data division, such as inconsistent predictions \cite{cream,trip} and uncertainty \cite{UGNCL}. To avoid inaccurate label predictions, some approaches \cite{l2rm,PC2} aim to refine alignments through alternative strategies like rematched mismatches \cite{l2rm} and pseudo captions \cite{PC2}. Besides these methods based data sanitized, other efforts \cite{decl,rcl,crcl} retort to robust loss functions to adaptively downweight the contributions of mismatches, e.g., evidential loss \cite{decl}, complementary contrast loss \cite{rcl}, and active complementary loss \cite{crcl}. Recently, some works \cite{esc,gsc} utilize the intrinsic properties observed within data to estimate accurate soft correspondence labels. However, thery all neglect the intra-modal relations, which is significantly crucial for accurately identify true correspondences.

\section{Methodology}

\subsection{Preliminaries}
\paragraph{Problem Definition}
In line with previous work, we take visual-text retrieval as a proxy task to discuss the noisy correspondence problem in cross-modal retrieval. Consider a multimodal dataset $\mathcal{D}=\left\{\left(V_{i}, L_{i}, y_{i}\right)\right\}_{i=1}^{N}$ containing of $N$ training pairs, where each $\left(V_{i}, L_{i}\right)$ denotes the $i$-th visual-text pair and $y_{i} \in \left\{0,1\right\}$ indicates whether the pair matched ($y_{i}=1$) or not ($y_{i}=0$). Typically, all pairs are assumed to be semantically associated with high similarity scores in the common representation space. However, due to the substantial costs of data collection and annotations, an unknown portion of mismatched pairs may be inadvertently labeled as matched ones. Such misalignment, a.k.a., noisy correspondence, without specific treatment, would severely disrupt the alignment between modalities and ultimately lead to performance degradation. The goal of our method is to effectively address the challenge of NCs within multimodal datasets, thus enabling robust cross-modal retrieval.

\paragraph{Intra-Modal Relation Alignment}
\label{sec:relation}
Given a sequence $\mathbf{O}=[o_{1}, \cdots, o_{N_{o}}]$ containing $N_{o}$ objects appeared in a visual-text pair, the sequences of visual and linguistic can be denoted as $\overline{\mathbf{V}}=[\overline{v}_{1},\cdots,\overline{v}_{N_{o}}]$ and $\overline{\mathbf{L}}=[\overline{l}_{1},\cdots,\overline{l}_{N_{o}}]$, respectively. Here, each item with same index corresponds to a same object. Note that an object may be described by one or more words in the sentence and one or more regions in the image, such that the linguistic item and visual item may represent a collocation of words and regions, respectively. The relation $\mathbf{C}_{o_{i}}=[c_{o_{i}\rightarrow o_{1}},\cdots,c_{o_{i}\rightarrow o_{N_{o}}}]$ of one object to others can also be depicted in both visual and linguistic modalities, i.e., $\mathbf{C}_{\overline{v}_{i}}=[c_{\overline{v}_{i}\rightarrow \overline{v}_{1}},\cdots,c_{\overline{v}_{i}\rightarrow \overline{v}_{N_{o}}}]$ and $\mathbf{C}_{\overline{l}_{i}}=[c_{\overline{l}_{i}\rightarrow \overline{l}_{1}},\cdots,c_{\overline{l}_{i}\rightarrow \overline{l}_{N_{o}}}]$, respectively.  Consequently, the alignment of such relations can be preserved by minimizing the expected risk for the distance objective~\cite{iais}, as expressed in the following equation:
\begin{equation}
    \mathcal{R}_{\mathcal{L}_{SD}} = \min\mathbb{E}_{(\mathbf{C}_{\overline{v}_{i}},\mathbf{C}_{\overline{l}_{i}}) \sim \mathcal{D}}[\mathcal{L}_{SD}(\mathbf{C}_{\overline{v}_{i}},\mathbf{C}_{\overline{l}_{i}})],
    \label{eq:relation}
\end{equation}
where $\mathcal{L}_{SD}$ is the loss function that utilized for narrowing semantic distance, e.g., symmetric matrix-based Kullback-Leibler Divergence (m-LK). Note that, IAIS \cite{iais} represents such relations within modalities using cross-modal attention matrix. Differently, ReCon obtains these relations by computing the similarity between objects within modalities.
 
\subsection{Relation Consistency Learning}
Let $\mathbf{V}{=}\left[v_{1},\cdots,v_{N_{\mathcal{V}}}\right]$ and $\mathbf{L}{=}\left[l_{1},\cdots,l_{N_{\mathcal{L}}}\right]$ be the original visual and linguistic input sequences, which respectively contains $N_{\mathcal{V}}$ visual regions and $N_{\mathcal{L}}$ linguistic words. The relation consistency learning aims not only to enforce alignment between objects across modalities, but also to ensure consistency of relations within modalities. Such dual constraints allow to comprehend more nuanced contextualized semantics compared to the relation-agnostic alignment and significantly improve the discriminability for true correspondences, which can effectively mitigate the risks of misleading caused by misidentified false supervisions, particularly in the presence of hard NCs. 

\paragraph{Cross-Modal Relation Consistency.} Cross-modal relation consistency refers to the semantic similarities between representations across modalities. To this end, two modal-specific networks $f_{\mathcal{V}}(\cdot, \Theta_{\mathcal{V}})$ and $f_{\mathcal{L}}(\cdot, \Theta_{\mathcal{L}})$ are first employed to project the visual and linguistic sequences into a common comparable space, where $\Theta_{\mathcal{V}}$ and $\Theta_{\mathcal{L}}$ are the parameterized models for visual and linguistic modalities, respectively. In the common space, the similarity of the given visual-linguistic pair is measured through similarity function $S=g(f_{\mathcal{V}}(\cdot),f_{\mathcal{L}}(\cdot),\Theta_{\mathcal{G}})$, where $\Theta_{\mathcal{G}}$ is the parameterized modal of similarity function $g$. Note that $g$ can be parametric \cite{sgraf, cmgm} or non-parametric \cite{gpo, vse++} function. For convenience, we denote $g(f_{\mathcal{V}}(\cdot),f_{\mathcal{L}}(\cdot))$ as $g(\cdot,\cdot)$ in the following. Intuitively, the goal of cross-modal consistency relation learning is to encourage the semantic gap between matches and mismatches as large as possible, which can be equivalent to maximizing the bidirectional matching probabilities of true correspondences. Consider a batch size $N_{b}$ pairs $\mathcal{D}_{N_{b}}=\{(V_{i}, L_{i}, y_{i})\}_{i=1}^{N_{b}}$, the matching probability of $i$-th visual query is defined as $p^{v2l}_{ij}=\frac{\exp(g\left(V_{i},L_{j}\right) / \tau)}{\sum_{k=1}^{N_{b}} \exp\left(g(V_{i},L_{k}\right) / \tau)}$, where $\tau$ is a temperature parameter. Likewise, the matching probability of $i$-th linguistic query is defined as $p^{l2v}_{ij}=\frac{\exp(g\left(V_{i},L_{j}\right) / \tau)}{\sum_{k=1}^{N_{b}} \exp(g\left(V_{k},L_{j}\right) / \tau)}$. Consequently, the cross-modal relation consistency can be preserved by minimizing the expected risk of bidirectional matching probabilities with the supervision of $y_{i}$, as expressed in the following equation:
\begin{equation}
    \mathcal{R}_{\mathcal{L}_{CM}} = \min \mathbb{E}_{(V_{i},L_{i},y_{i} ) \sim \mathcal{D}_{N_{b}}} [\mathcal{L}_{CM}\left(V_{i},L_{i},y_{i}\right)],
\end{equation}
where $\mathcal{L}_{CM}$ denotes the cross-modal InfoNCE loss \cite{clip}, which encourages the similarity gap between matched and mismatched pairs as large as possible. Note that, the contributions of different data pairs will be adaptively adjusted according to their corresponding supervisions $y$. 

\paragraph{Intra-Modal Relation Consistency.}
Intra-modal relation consistency refers to the matching of semantics between visual contexts among regions and linguistic contexts among words. Unfortunately, the absence of explicit object annotations presents a particularly intricate and demanding challenge, which is quite common in real-world scenarios. Consequently, we cannot access to the sequences containing objects, which means that each visual/linguistic item now corresponds to only one region/word. Undoubtedly, Eq. \eqref{eq:relation} cannot be directly applied to such input sequences due to the lack of one-to-one correspondence properties found in object sequences. To address this problem, like~\cite{iais}, we first select an anchor modality, e.g., visual modality, containing regions sequence, and then construct a proxy sequence containing the most corresponding words sequence from the opposite modality, such that the relations of distinct modalities can be comparable. Gven a sequence $V$ with visual modality as anchor, the relations to sequence $L$ from the opposite modality can be obtained by $\mathbf{C}_{\mathcal{VL}}=g(V,L)$, wherein the relations between every visual item $v_{i}$ to all the linguistic items are depicted in $\mathbf{C}_{\mathcal{VL}}[i,:]$, i.e., the $i$-th row of this relation matrix. Subsequently, we can obtain the most relevant item $l_{i^*}$ for $v_{i}$, wherein the index can be calculated as $i^*=\arg \max \mathbf{C}_{\mathcal{VL}}[i,:]$. Likewise, we can obtain the most relevant linguistic item $l_{j^*}$ for the visual item $v_{j}$. Therefore, the intra-modal relations $c_{v_{i} \rightarrow v_{j}}$ within visual modality can be depicted by the intra-modal relations $c_{l_{i^*} \rightarrow l_{j^*}}$ within linguistic modality, which can be formulated in the following equation:
\begin{equation}
    \mathbf{C}_{\mathcal{VV}}^{p} = \{c^{p}_{v_{i} \rightarrow v_{j}} | i,j\in[1,N_{\mathcal{V}}] \} = \Psi(\mathbf{C}_{\mathcal{LL}},l_{i^*},l_{j^*}),
    \label{eq:proxy}
\end{equation}
where $\Psi$ represents a reconstruction operation that form the proxy relation matrix. Here, the $\mathbf{C}_{\mathcal{VV}}^{p}$ can be regarded as a representation of the original visual relation matrix $\mathbf{C}_{\mathcal{VV}}$ from the linguistic view. Similarly, with linguistic modality as anchor, the reconstructed proxy relation matrix $\mathbf{C}_{\mathcal{LL}}^{p}$ from the visual view, which depicts the relations within linguistic modality, can also be obtained through above Eq. \eqref{eq:proxy}. As discussed in Sec. \ref{sec:relation}, we can now employ the m-LK to compute the distances between relation matrix and its proxy relation matrix, which can be defined as:
\begin{equation}
    \mathcal{L}_{IM} = D_{KL}(\mathbf{C}_{\mathcal{VV}}||\mathbf{C}_{\mathcal{VV}}^{p}) + D_{KL}(\mathbf{C}_{\mathcal{LL}}||\mathbf{C}_{\mathcal{LL}}^{p}).
\end{equation}

Therefore, the preservation of relations within modalities can be achieved through minimizing the expected risk of the above $\mathcal{L}_{IM}$, as formulated in the following equation:
\begin{equation}
    \mathcal{R}_{\mathcal{L}_{IM}} = \min \mathbb{E}_{(V_{i},L_{i},y_{i}) \sim \mathcal{D}_{N_{b}}}[ \mathcal{L}_{IM}(V_{i},L_{i},y_{i})].
\end{equation}

\subsection{True Correspondence Discrimination}
Due to the existence of NC, we can only have access to the noisy training dataset $\Tilde{\mathcal{D}}$ containing an unknown proportion of mismatched pairs. Thus, directly optimizing models on such dataset using the above loss functions may risks misleading by the unwanted mismatched pairs, potentially causing significant performance degradation or even leading to training collapse. To address this problem, a common strategy \cite{ncr, BiCro, esc} is to leverage the small-loss criterion \cite{dividemix} to divide the noisy dataset into clean and noisy partitions, wherein the different partitions will be processed with corresponding training strategies. In details, the clean partition can be directly used for model optimization, while the corresponding strategy might be exploited to learn all available and informative knowledge from the noisy partition, e.g., locally-associated correspondences, avoiding insufficient utilization for dataset. However, the previous methods may misidentifies the mismatched pairs as matches, thus declining the discriminability for true correspondences and resulting in suboptimal performance due to the misleading of mismatches. Thanks to the dual constrains of relations, ReCon provides more refined and reliable data division and effectively mitigates the misidentification of mismatches, especially in the existence of hard NCs.

\paragraph{Noisy Data Division.}
Inspired by the previous success \cite{ncr,BiCro}, we also leverage the small-loss criterion to achieve a rough division for the corrupted training data. Specifically, we first compute the per-sample cross-modal relation loss by $\mathcal{L}_{CM}$, denoted as $\{l_{i}^{CM}\}_{i=1}^{N}=\{\mathcal{L}_{CM}(V_{i},L_{i})\}_{i=1}^{N}$. Next, a two-component Gaussian Mixture Model (GMM) \cite{gmm} would be employed to fit the per-sample loss distribution of all training pairs, which can be expressed as $p(l_{i}^{CM})=\sum_{k=1}^{K}\lambda_{k}\phi(l_{i}^{CM} | k)$. Here $K=2$, $\lambda_{k}$ is the corresponding mixture coefficient, and $\phi(l_{i}^{CM} | k)$ indicates the probability density function of the $k$-th component. Besides, the Expectation-Maximization algorithm is employed to optimize the GMM. Finally, we use the component with smaller mean to obtain the estimated probability:
\begin{equation}
    y_{i}^{CM}=p(k|l_{i}^{CM})=p(k)p(l_{i}^{CM}|k) / p(l_{i}^{CM}).
\end{equation}

By setting a threshold $\omega_{1}$, we can roughly divide the dataset $\tilde{\mathcal{D}}$ into rough clean partition $\tilde{\mathcal{D}}_{c}=\{(V_{i},L_{i})|y_{i}^{CM} > \omega_{1}\}$ and noisy partition $\mathcal{D}_{n}=\{(V_{i},L_{i})|y_{i}^{CM} \leq \omega_{1}\}$. Theoretically, the probability of positive pairs should approach 1, while for negative pairs, it should tend toward 0. 

\paragraph{True Positives Identification.}
To ensure that the model learns accurate representations of matched data pairs and their relations, establishing a reliable division for true positives is crucial. In practice, the accurate discrimination for positives contributes more than negatives, for only true positives can effectively guide model optimization and further enhance its discriminability, thus minimizing the risk caused by wrong supervisions. Even if some positive pairs are wrongly divided into the noisy partition, they can still be learned through the corresponding strategy of handling the noisy partition. However, the misidentified negatives would directly compromise the discriminability of model, which further increase the risk of learning from the false correspondences. Consequently, we employ the cross-modal and intra-modal relation consistency to jointly discriminate the true positives, and the discrepancies of relations within modalities can be measured through the following equation:
\begin{equation}
    y_{i}^{IM} = \frac{ \log(1 + \mathcal{L}_{IM}(V_{i},L_{i}))}{1 + \log(1 + \mathcal{L}_{IM}(V_{i},L_{i}))}.
\end{equation}

Theoretically, the discrepancies for true correspondences should approach zero, while others will exhibit significantly larger discrepancies due to their inconsistent intra-modal relations. Thus, we can distinguish such pairs from the $\tilde{\mathcal{D}}_{c}$ by a fixed threshold $\omega_{2}$ to form two refined partitions:
\begin{equation}
    \begin{cases}
        \mathcal{D}_{c} = \{(V_{i},L_{i})|y_{i}^{IM} < \omega_{2}, \forall(V_{i},L_{i}) \in \tilde{\mathcal{D}}_{c} \} \\
        \mathcal{D}_{l} = \{(V_{i},L_{i})|y_{i}^{IM} \geq \omega_{2}, \forall(V_{i},L_{i}) \in \tilde{\mathcal{D}}_{c} \}
    \end{cases},
\end{equation}
where $\mathcal{D}_{c}$ denotes the clean partition containing true correspondences that can be directly employed to subsequent training and $\mathcal{D}_{l}$ contains pairs of local-associated correspondences. To fully learn all available local-associated correspondences and  enhance the discriminability of models, while simultaneously enlarge the semantic distance between true correspondences and others, we penalize the weight of pairs belonging to $\mathcal{D}_{l}$ based on their discrepancies of intra-modal relations. The specific penalization factor can be calculated as follows:
\begin{equation}
\label{penalization}
    \lambda = \exp\{y_{i}^{IM} / \alpha\},
\end{equation}
where $\alpha$ is an empirical scale parameter. For the pairs belonging to $\mathcal{D}_{n}$, we estimate pseudo labels through the predictions of models to replace the original unreliable labels, which can be expressed as:
\begin{equation}
    \tilde{y}_{i}^{t} = \beta \tilde{y}_{i}^{t-1} + (1-\beta)p^{t}(V_{i},L_{i}), \forall (V_{i},L_{i}) \in \mathcal{D}_{n},
\end{equation}
where $\beta$ is the momentum coefficient, $\tilde{y}_{i}^{t}$ represents the estimated labels at $t$-th epoch and $p(V_{i},L_{i})=(p_{ii}^{v2t} + p_{ii}^{t2v}) / 2$ denotes the average matching probability. Thus, the final recasted labels of all pairs can be summarized as follows:
\begin{equation}
    \hat{y}_{i} = \begin{cases}
        1, \forall \left(V_{i},L_{i} \right) \in \mathcal{D}_{c} \cup \mathcal{D}_{l} \\
        \tilde{y}_{i}, \forall \left(V_{i},L_{i} \right) \in \mathcal{D}_{n}
    \end{cases}.
\end{equation}

\subsection{Overall Optimization Objective}
To ensure the initial stability and convergence for subsequent training, we first conduct $\eta$ epochs warmup process using the triplet loss \cite{vsrn}, which can be denoted as follows:
\begin{equation}
    \begin{aligned}
        \mathcal{L}_{w} &= \sum_{\tilde{L}}[\gamma - g(V_{i},L_{i}) + g(V_{i},\tilde{L})]_{+} \\
    &+ \sum_{\tilde{V}}[\gamma - g(V_{i},L_{i}) + g(\tilde{V},L_{i})]_{+}
    \end{aligned},
\end{equation}
where $\gamma$ is the fixed margin that controls the distance between positives and negatives, $[x]_{+}=\max(x,0)$, $\tilde{L}$ and $\tilde{V}$ are the negative samples in a given mini batch. Afterwards, the different partitions will be trained with corresponding optimization strategies. For pairs belonging to the $\mathcal{D}_{c}$, we aim to learn correct representations of matched pairs and relations, thus enhancing the discriminability for true correspondences.
Hence, the loss function for $\mathcal{D}_{c}$ is a combination of $\mathcal{L}_{CM}$ and $\mathcal{L}_{IM}$:
\begin{equation}
    \mathcal{L}_{c} = \xi \mathcal{L}_{CM} + \mathcal{L}_{IM},
\end{equation}
where $\xi$ is the balance factor that adjusts the contributions of cross-modal relations. As for the pairs belonging to $\mathcal{D}_{l}$, the penalization factor calculated by Eq. \eqref{penalization} will be employed to downweight the contributions of intra-modal relations:
\begin{equation}
    \mathcal{L}_{l} = \xi \mathcal{L}_{CM} + \frac{1}{\lambda} \mathcal{L}_{IM}.
\end{equation}

Finally, for the pairs belonging to $\mathcal{D}_{n}$, the estimated pseudo labels will be employed to adjust their contributions in cross-modal relations, while the intra-modal relations will be excluded to avoid incorrect supervisions:
\begin{equation}
    \mathcal{L}_{n} = \hat{y}_{i}\mathcal{L}_{CM} = \mathcal{H}(\hat{y}_{i},p_{ii}^{v2l}) + \mathcal{H}(\hat{y}_{i},p_{ii}^{l2v}),
\end{equation}
where $\mathcal{H}$ denotes the batched cross-entropy function.

\section{Experiments}

\subsection{Datasets and Protocols}
\paragraph{Datasets.}
We evaluate our method on three widely-used benchmarks, following the settings in \cite{ncr}. Specifically, Flickr30K \cite{flickr} contains 31K images with five textual descriptions, collected from the Flickr website. We split 1K image-text pairs for validation, 1K pairs for testing, and the rest are assigned for training. MS-COCO \cite{coco} includes 123, 287 images with five associated captions each. We assign 113, 287 image-text pairs for model training, 5K pairs for validation, and the rest for testing. Both results are reported in our experiments by averaging over 5 folds of 1K test pairs and on the whole 5K test pairs. Conceptual Captions (CC) \cite{cc} is a web-crawled large-scale dataset automatically sourced from the Internet, which inadvertently contains about 3\%$\sim$20\% mismatched or weakly-matched pairs, i.e., noisy correspondence. In our experiments, CC152K, a subset of CC, is utilized for model evaluation, which comprises 1K image-text pairs designated for validation, 1K pairs for testing, and the remaining 150K pairs for training.

\paragraph{Evaluation Protocols.}
Recall at K (R@K) is a widely-used metric to measure the retrieval performance, defined as the percentage of matched items successfully retrieved from the top K candidates~\cite{PAMI2021}. In our experiments, the R@1, R@5, R@10, and the sum of three recalls for image-to-text and text-to-image retrieval are all reported to provide a comprehensive performance evaluation for our method.

\begin{table}
    \centering
    \caption{Comparisons with real-world NCs on CC152K. The \textbf{Best} and \underline{second-best} results are respectively marked in each column.}
    \label{tab:real_world}
    \setlength{\tabcolsep}{0.98mm}
    \begin{tabular}{c|ccccccc}
    \toprule
        \multirow{2}{*}{Methods} & \multicolumn{3}{c}{Image to Text} & \multicolumn{3}{c}{Text to Image} & \\
         \cline{2-8}
        & R@1 & R@5 &R@10 & R@1 & R@5 &R@10 & rSum \\
       \bottomrule
       SCAN & 30.5 & 55.3 & 65.3 & 26.9 & 53.0 & 64.7 & 295.7 \\
       NCR & 39.5 & 64.5 & 73.5 & 40.3 & 64.6 & 73.2 & 355.6 \\
       DECL & 39.0 & 66.1 & 75.5 & 40.7 & 66.3 & 76.7 & 364.3 \\
       MSCN & 40.1 & 65.7 & 76.6 & 40.6 & 67.4 & 76.3 & 366.7 \\
       BiCro & 40.8 & 67.2 & 76.1 & 42.1 & 67.6 & 76.4 & 370.2 \\
       RCL & 41.7 & 66.0 & 73.6 & 41.6 & 66.4 & 75.1 & 364.4 \\
       CRCL & 41.8 & 67.4 & 76.5 & 41.6 & 68.0 & \textbf{78.4} & 373.7 \\
       SREM & 40.9 & 67.5 & 77.1 & 41.5 & \underline{68.2} & 77.0 & 372.2 \\
       PC$^2$ & 39.3 & 66.4 & 75.4 & 39.8 & 66.4 & 76.8 & 364.1 \\
       L2RM & \underline{43.0} & 67.5 & 75.7 & 42.8 & 68.0 & 77.2 & 374.2 \\
       ESC & 42.8 & 67.3 & 76.9 & \underline{44.8} & \underline{68.2} & 75.9 & \underline{375.9} \\
       GSC & 42.1 & \underline{68.4} & \underline{77.7} & 42.2 & 67.6 & 77.1 & 375.1 \\
       \textbf{ReCon} & \textbf{43.1} & \textbf{68.7} & \textbf{78.1} & \textbf{44.9} & \textbf{68.3} & \underline{77.4} & \textbf{380.5} \\
       \bottomrule
    \end{tabular}
\end{table}

\subsection{Implementation Details}
For fair comparisons, all experiments are conducted using the same backbone SGRAF \cite{sgraf} and all experimental settings are consistent with NCR \cite{ncr}, except for the specific parameters of ReCon. Specifically, the batch size $N_{b}$ is set to 128 and the temperature coefficients $\tau$ is 0.1. The division thresholds $\omega_{1}$ and $\omega_{2}$ are both set to 0.5, the scale parameter $\alpha$ for penalization factor is set to 0.1, and the momentum coefficient $\beta$ is 0.6. Moreover, the fixed margin $\gamma$ is set to 0.2 and the balance factor $\xi$ is 5. Before training models, we conduct a $\eta=5$ epochs warmup process for initial convergence. Besides, all experiments are conducted without any additional preprocessing or the use of external data sources.


\begin{table*}
    \centering
    \setlength{\tabcolsep}{0.9mm}
    \caption{Cross-modal retrieval performance comparison under synthetic noise rates of 20\%, 40\%, and 60\% on Flickr30K and MS-COCO 1K. The best and the second best results are respectively marked by \textbf{bold} and \underline{underline}.}
    \label{tab:simulated}
    \begin{tabular}{c|c|ccc|ccc|c|ccc|ccc|c}
        \toprule
        \multirow{3}{*}{\makecell[c]{Noise \\ Ratio}} & \multirow{3}{*}{Methods} & \multicolumn{7}{|c}{Flickr30K} & \multicolumn{7}{|c}{MS-COCO 1K} \\
        & & \multicolumn{3}{|c|}{Image to Text} & \multicolumn{3}{|c|}{Text to Image} & & \multicolumn{3}{|c|}{Image to Text} & \multicolumn{3}{|c|}{Text to Image} & \\
        \cline{3-16}
         &  & R@1 & R@5 & R@10 & R@1 & R@5 & R@10 & rSum & R@1 & R@5 & R@10 & R@1 & R@5 & R@10 & rSum\\
        \bottomrule
        \multirow{13}{*}{20\%} 
        & SCAN (ECCV'18) & 59.1 & 83.4 & 90.4 & 36.6 & 67.0 & 77.5 & 414.0 & 66.2 & 91.0 & 96.4 & 45.0 & 80.2 & 89.3 & 468.1 \\
        & NCR (NIPS'21) & 73.5 & 93.2 & 96.6 & 56.9 & 82.4 & 88.5 & 491.1 & 76.6 & 95.6 & 98.2 & 60.8 & 88.8 & 95.0 & 515.0 \\
        & DECL (ACM MM'22) & 77.5 & 93.8 & 97.0 & 56.1 & 81.8 & 88.5 & 494.7 & 77.5 & 95.9 & 98.4 & 61.7 & 89.3 & 95.4 & 518.2 \\
        & MSCN (CVPR'23) & 77.4 & 94.9 & 97.6 & 59.6 & 83.2 & 89.2 & 501.9 & 78.1 & \textbf{97.2} & 98.8 & 64.3 & 90.4 & 95.8 & 524.6 \\
        & BiCro (CVPR'23) & 78.1 & 94.4 & 97.5 & 60.4 & 84.4 & 89.9 & 504.7 & 78.8 & 96.1 & 98.6 & 63.7 & 90.3 & 95.7 & 523.2 \\
        & RCL (TPAMI'23) & 75.9 & 94.5 & 97.3 & 57.9 & 82.6 & 88.6 & 496.8 & 78.9 & 96.0 & 98.4 & 62.8 & 89.9 & 95.4 & 521.4 \\
        & CRCL (NIPS'23) & 78.9 & 94.8 & \textbf{97.9} & 58.7 & 83.0 & 89.2 & 502.5 & 77.8 & 96.1 & 98.5 & 63.4 & 90.3 & \underline{95.9} & 522.0 \\
        & SREM (AAAI'24) & \underline{79.5} & 94.2 & \textbf{97.9} & \underline{61.2} & \underline{84.8} & 90.2 & \underline{507.8} & 78.5 & 96.8 & 98.8 & 63.8 & 90.4 & 95.8 & 524.1 \\
        & PC$^2$ (ACM MM'24) & 78.7 & 94.9 & 96.9 & 59.8 & 83.9 & 89.6 & 503.8 & 77.8 & 95.7 & 98.4 & 62.8 & 89.7 & 95.3 & 519.7 \\
        & L2RM (CVPR'24) & 77.9 & \underline{95.2} & \underline{97.8} & 59.8 & 83.6 & 89.5 & 503.8 & \underline{80.2} & 96.3 & 98.5 & 64.2 & 90.1 & 95.4 & 524.7 \\
        & ESC (CVPR'24) & 79.0 & 94.8 & 97.5 & 59.1 & 83.8 & 89.1 & 503.3 & 79.2 & \underline{97.0} & \textbf{99.1} & \underline{64.8} & \underline{90.7} & \textbf{96.0} & \underline{526.8} \\
        & GSC (CVPR'24) & 78.3 & 94.6 & \underline{97.8} & 60.1 & 84.5 & \underline{90.5} & 505.8 & 79.5 & 96.4 & \underline{98.9} & 64.4 & 90.6 & \underline{95.9} & 525.7 \\
        
        & \textbf{ReCon} & \textbf{80.3} & \textbf{95.3} & \underline{97.8} & \textbf{61.6} & \textbf{85.5} & \textbf{91.3} & \textbf{511.8} & \textbf{80.9} & 96.6 & 98.8 & \textbf{65.2} & \textbf{91.0} & \textbf{96.0} & \textbf{528.6} \\
        \bottomrule
        
        \multirow{13}{*}{40\%}
        & SCAN (ECCV'18) & 29.9 & 60.5 & 72.5 & 16.4 & 38.5 & 48.6 & 266.4 & 30.1 & 65.2 & 79.2 & 18.9 & 51.1 & 69.9 & 314.4 \\
        & NCR (NIPS'21) & 75.3 & 92.1 & 95.2 & 56.2 & 80.6 & 87.4 & 486.8 & 76.5 & 95.0 & 98.2 & 60.7 & 88.5 & 95.0 & 513.9 \\
        & DECL (ACM MM'22) & 72.7 & 92.3 & 95.4 & 53.4 & 79.4 & 86.4 & 479.6 & 75.6 & 95.5 & 98.3 & 59.5 & 88.3 & 94.8 & 512.0 \\
        & MSCN (CVPR'23) & 74.4 & \textbf{94.4} & \underline{96.9} & 57.2 & 81.7 & 87.6 & 492.2 & 74.8 & 94.9 & 98.0 & 60.3 & 88.5 & 94.4 & 510.9 \\
        & BiCro (CVPR'23) & 74.6 & 92.7 & 96.2 & 55.5 & 81.1 & 87.4 & 487.5 & 77.0 & 95.9 & 98.3 & 61.8 & 89.2 & 94.9 & 517.1 \\
        & RCL (TPAMI'23) & 72.7 & 92.7 & 96.1 & 54.8 & 80.0 & 87.1 & 483.4 & 77.0 & 95.5 & 98.3 & 61.2 & 88.5 & 94.8 & 515.3 \\
        & CRCL (NIPS'23) & 74.1 & 92.6 & \underline{96.9} & 55.5 & 80.9 & 87.6 & 487.6 & 76.6 & 95.6 & 98.5 & 62.3 & 89.7 & \underline{95.4} & 518.1 \\
        & SREM (AAAI'24) & \underline{76.5} & 93.9 & 96.3 & \underline{57.5} & \underline{82.7} & 88.5 & 495.4 & 77.2 & 96.0 & 98.5 & 62.1 & 89.3 & 95.3 & 518.4 \\
        & PC$^2$ (ACM MM'24) & 75.8 & 93.5 & \underline{96.9} & \underline{57.5} & 81.9 & 88.2 & 493.8 & 77.4 & 95.8 & 98.4 & 62.1 & 89.4 & 95.1 & 518.2 \\
        & L2RM (CVPR'24) & 75.8 & 93.2 & \underline{96.9} & 56.3 & 81.0 & 87.3 & 490.5 & 77.5 & 95.8 & 98.4 & 62.0 & 89.1 & 94.9 & 517.7\\
        & ESC (CVPR'24) & 76.1 & 93.1 & 96.4 & 56.0 & 80.8 & 87.2 & 489.6 & \underline{78.6} & \textbf{96.6} & \textbf{99.0} & \underline{63.2} & \textbf{90.6} & \textbf{95.9} & \underline{523.9} \\
        & GSC (CVPR'24) & \underline{76.5} & 94.1 & \textbf{97.6} & \underline{57.5} & \underline{82.7} & \underline{88.9} & \underline{497.3} & 78.2 & 95.9 & 98.2 & 62.5 & 89.7 & \underline{95.4} & 519.9 \\
        
        & \textbf{ReCon} & \textbf{79.4} & \underline{94.3} & \textbf{97.6} & \textbf{59.9} & \textbf{83.9} & \textbf{90.1} & \textbf{505.2} & \textbf{79.9} & \underline{96.2} & \underline{98.6} & \textbf{63.5} & \underline{90.5} & \textbf{95.9} & \textbf{524.5} \\
        \bottomrule
        
        \multirow{13}{*}{60\%}
        & SCAN (ECCV'18) & 16.9 & 39.3 & 53.9 & 2.8 & 7.4 & 11.4 & 131.7 & 27.8 & 59.8 & 74.8 & 16.8 & 47.8 & 66.4 & 293.4 \\
        & NCR (NIPS'21) & 68.7 & 89.9 & 95.5 & 52.0 & 77.6 & 84.9 & 468.6 & 72.7 & 94.0 & 97.6 & 57.9 & 87.0 & 94.1 & 503.3 \\
        & DECL (ACM MM'22) & 65.2 & 88.4 & 94.0 & 46.8 & 74.0 & 82.2 & 450.6 & 73.0 & 94.2 & 97.9 & 57.0 & 86.6 & 93.8 & 502.5 \\
        & MSCN (CVPR'23) & 70.4 & 91.0 & 94.9 & 53.4 & 77.8 & 84.1 & 471.6 & 74.4 & \underline{95.1} & 97.9 & 59.2 & 87.1 & 92.8 & 506.5 \\
        & BiCro (CVPR'23) & 67.6 & 90.8 & 94.4 & 51.2 & 77.6 & 84.7 & 466.3 & 73.9 & 94.4 & 97.8 & 58.3 & 87.2 & 93.9 & 505.5 \\
        & RCL (TPAMI'23) & 67.7 & 89.1 & 93.6 & 48.0 & 74.9 & 83.3 & 456.6 & 74.0 & 94.3 & 97.5 & 57.6 & 86.4 & 93.5 & 503.3 \\
        & CRCL (NIPS'23) & 70.4 & 90.4 & 94.9 & 52.6 & 78.1 & 85.1 & 471.5 & 75.2 & 94.9 & 98.0 & 60.1 & 88.5 & 94.8 & 511.5 \\
        & SREM (AAAI'24) & 71.0 & \underline{92.1} & \underline{96.1} & \underline{54.0} & \underline{80.1} & \underline{87.0} & \underline{480.3} & 74.5 & 94.5 & 97.9 & 58.7 & 87.5 & 93.9 & 506.9 \\
        & PC$^2$ (ACM MM'24) & 70.8 & 90.3 & 94.4 & 53.1 & 79.0 & 85.9 & 473.5 & 74.2 & 94.4 & 97.8 & 58.9 & 87.5 & 93.8 & 506.6 \\
        & L2RM (CVPR'24) & 70.0 & 90.8 & 95.4 & 51.3 & 76.4 & 83.7 & 467.6 & 75.4 & 94.7 & 97.9 & 59.2 & 87.4 & 93.8 & 508.4 \\
        & ESC (CVPR'24) & \underline{72.6} & 90.9 & 94.6 & 53.0 & 78.6 & 85.3 & 475.0 & \textbf{77.2} & \underline{95.1} & \underline{98.1} & \underline{61.1} & \underline{88.6} & \underline{94.9} & \underline{515.0} \\
        & GSC (CVPR'24) & 70.8 & 91.1 & 95.9 & 53.6 & 79.8 & 86.8 & 478.0 & \underline{75.6} & \underline{95.1} & 98.0 & 60.0 & 88.3 & 94.6 & 511.7 \\

        & \textbf{ReCon} & \textbf{74.3} & \textbf{93.6} & \textbf{96.6} & \textbf{55.7} & \textbf{81.6} & \textbf{88.1} & \textbf{489.9} & \textbf{77.2} & \textbf{95.9} & \textbf{98.4} & \textbf{61.8} & \textbf{89.3} & \textbf{95.2} & \textbf{517.8} \\
        \bottomrule
    \end{tabular}
\end{table*}

\subsection{Comparison with State-of-the-Arts}
In this section, we carry out a comprehensive evaluation to present the effectiveness of ReCon, benchmarking it against SOTA baselines across three widely-used datasets above. The baselines comprise SCAN \cite{scan}, NCR \cite{ncr}, DECL \cite{decl}, MSCN \cite{MSCN}, BiCro \cite{BiCro}, RCL \cite{rcl}, CRCL \cite{crcl}, SREM \cite{srem}, PC$^2$ \cite{PC2}, L2RM \cite{l2rm}, ESC \cite{esc} and GSC \cite{gsc}. For the well-established Flickr30K and MS-COCO, the simulated NCs with varying noise rates, namely 20\%, 40\%, and 60\%, obtained by randomly shuffling the captions like \cite{ncr} are exploited to assess the robustness of ReCon. In addition to simulated NCs, we also validate the performance of ReCon with real-world noisy conditions using the web-crawled CC152K naturally containing 3\% $\sim$ 20\% unknown NCs. Note that the presented results of ReCon on the testing set are obtained through the checkpoints that achieved optimal performance on the validation set.

\textbf{Results on Simulated NCs.}
For quantitative evaluation the performance and robustness of all baselines under different noise ratios, we conduct all tested baselines on the Flickr30K and MS-COCO 1K with 20\%, 40\%, and 60\% of simulated noisy correspondence, where the results of MS-COCO are averaged on 5 folds of 1K test pairs as in previous works \cite{ncr, crcl}. The details are recorded in Table \ref{tab:simulated}, which demonstrates that our ReCon remarkably outperforms other baselines by a large margin on most of metrics. Notably, ReCon gains the highest R@1 score for both image-to-text and text-to-image retrieval across all noise rates on these two datasets, indicating that our method has significant potential to effectively deal with NCs. This promising performance can be attributed to the accurate identification for true positives, which avoids the misleading of wrongly introduced mismatches and enhances the discrimination between matched and mismatched pairs, thus achieving further performance improvement. Besides, ReCon performs competitive performance than other baselines under severely noise, proving its stability and reliability to facilitate robust cross-modal retrieval.


\begin{table}
    \centering
    \caption{Performance comparison with CLIP on MS-COCO 5K. The \textbf{best} results are highlighted in \textbf{bold}.}
    \label{clip}
      \setlength{\tabcolsep}{0.5mm}
    \begin{tabular}{c|c|ccccccc}
    \toprule
        \multirow{2}{*}{Noise} & \multirow{2}{*}{Methods} & \multicolumn{3}{c}{Image to Text} & \multicolumn{3}{c}{Text to Image} & \\
    \cline{3-9}
         &  & R@1 & R@5 &R@10 & R@1 & R@5 &R@10 & rSum \\
    \midrule
        \multirow{3}{*}{0\%} & CLIP-14 & 58.4 & 81.5 & 88.1 & 37.8 & 62.4 & 72.2 & 400.4 \\
         & CLIP-32 & 50.2 & 74.6 & 83.6 & 30.4 & 56.0 & 66.8 & 361.6 \\
         & \textbf{ReCon} & \textbf{61.6} & \textbf{86.7} & \textbf{92.7} & \textbf{44.4} & \textbf{73.1} & \textbf{83.1} & \textbf{441.6} \\
    \midrule
        \multirow{3}{*}{20\%} & CLIP-14 & 36.1 & 61.3 & 72.5 & 22.6 & 43.2 & 53.7 & 289.4 \\
         & CLIP-32 & 21.4 & 49.6 & 63.3 & 14.8 & 37.6 & 49.6 & 236.3 \\
         & \textbf{ReCon} & \textbf{61.1} & \textbf{85.7} & \textbf{92.2} & \textbf{43.5} & \textbf{72.4} & \textbf{82.7} & \textbf{437.6} \\
    \midrule
        \multirow{2}{*}{50\%} & CLIP-32 & 10.9 & 27.8 & 38.3 & 7.8 & 19.5 & 26.8 & 131.1 \\
         & \textbf{ReCon} & \textbf{58.1} & \textbf{85.1} & \textbf{91.9} & \textbf{41.5} & \textbf{70.7} & \textbf{81.0} & \textbf{428.3} \\
    \bottomrule
    \end{tabular}
\end{table}

\begin{figure}
    \centering
        \begin{minipage}{0.25\textwidth}
            \centering
            \includegraphics[width=\textwidth]{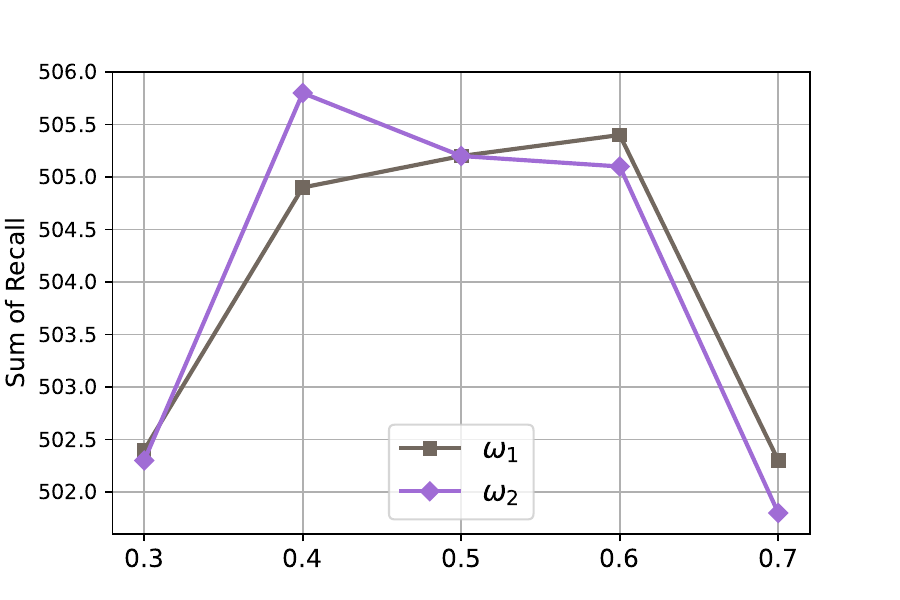}
        \end{minipage}
        \begin{minipage}{0.21\textwidth}
            \centering
            \includegraphics[width=\textwidth]{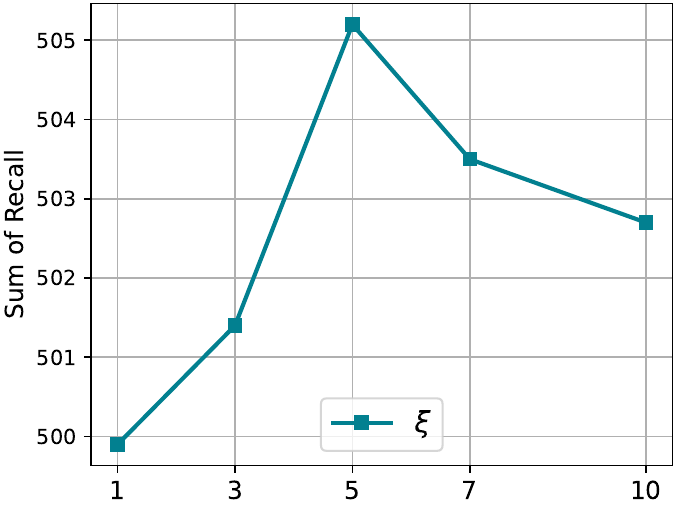}
        \end{minipage}
    \caption{Performance under different hyper-parameters of ReCon on Flickr30K with 40\% NCs.}
    \label{fig:hyper}
\end{figure}

\textbf{Results on Real-World NCs.}
For substantiating the comprehensive performance assessment, we also provide the quantitative results that evaluated on CC152K containing real-world NCs, which better mirrors real-world industry scenarios. According to the results shown in Table \ref{tab:real_world}, it can be observed that ReCon outperforms the baselines by a considerable margin with the overall score 4.4\% performance improvement compared to the second-best ESC of 375.9\%. Besides, ReCon exhibits competitive performance across all metrics, consistently indicating its robustness and effectiveness in handling real-world NCs.

\textbf{Comparison to Pre-trained Model.}
To further present the superiority and necessity of ReCon, we perform comparisons to the large pre-trained vision-language model, i.e., CLIP \cite{clip}, which is a powerful baseline trained on massive image-text pairs collected from the Internet with a large number of real NCs. In line with \cite{ncr}, we compare our ReCon to the CLIP on MS-COCO dataset under the following two settings: zero-shot and fine-tune, and the two baselines: CLIP-14 (ViT-L/14) and CLIP-32 (ViT-B/32). From the results shown in Table \ref{clip}, the significant performance degradation of CLIP can be attribute to the lack of effective mechanism to handle noisy correspondence. In contrast, the performance of ReCon under 50\% noise even surpasses the zero-shot results achieved of CLIP, indicating the effectiveness and necessity of our ReCon.

\begin{table}
    \centering
      \caption{Ablation studies on Flick30K with 40\% noise with different components in ReCon. The \textbf{best} results are marked in \textbf{bold}.}
    \label{tab:ablation}
    \setlength{\tabcolsep}{0.8mm}
    \begin{tabular}{ccc|ccccccc}
    \toprule
        \multicolumn{3}{c}{Components} & \multicolumn{3}{|c}{Image to Text} & \multicolumn{3}{|c}{Text to Image} &  \\
        \midrule
        Tru. & $\mathcal{L}_{IM}$ & $\lambda$ & R@1 & R@5 &R@10 & R@1 & R@5 &R@10 & rSum \\
        \midrule
        \checkmark & \checkmark & \checkmark & \textbf{79.4} & \textbf{94.3} & \textbf{97.6} & \textbf{59.9} & \textbf{83.9} & \textbf{90.1} & \textbf{505.2} \\
        \checkmark & \checkmark &  & 77.3 & 94.1 & 97.3 & 58.7 & 83.3 & 89.5 & 500.2 \\
        & \checkmark & \checkmark & 77.2 & \textbf{94.3} & 97.2 & 57.9 & 83.1 & 89.3 & 499.1 \\
        & \checkmark &  & 77.0 & 94.1 & 97.0 & 57.6 & 82.8 & 89.0 & 497.5 \\
        \checkmark & & & 74.1 & 93.2 & 96.7 & 57.4 & 83.1 & 88.9 & 493.3 \\
         \bottomrule
    \end{tabular}
\end{table}

\begin{figure}
    \centering
    \includegraphics[width=1.0\linewidth]{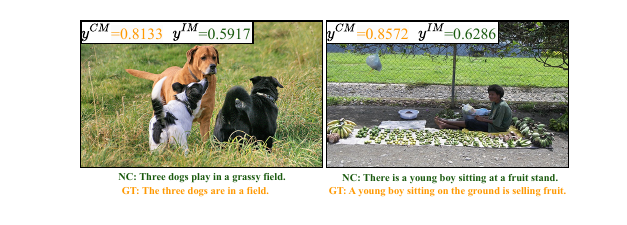}
    \caption{Examples of detected mismatched pairs on Flickr30K.}
    \label{fig:case}
\end{figure}

\subsection{Ablation Study}
\textbf{Impact of components.}
We conducted ablation studies on the Flickr30K with 40\% noise to validate the individual contributions of each component within ReCon, as detailed in Table \ref{tab:ablation}. For the true correspondence discrimination, all pairs are divided into clean and noisy partitions based on the $\mathcal{L}_{CM}$, and the intra-modal relation $\mathcal{L}_{IM}$ is directly employed to the clean partition. From the table, ReCon achieves the optimal performance by integrating all these components. This substantial improvement not only confirms the effectiveness of each individual component but also indicates their collective contributions in enhancing the robustness of models to address noisy correspondence.
\textbf{Impact of hyper-parameters.}
Fig. \ref{fig:hyper} shows the effects of the main hyper-parameters including division thresholds and balance factor. From the results, ReCon obtains better performance with $\omega_{1},\omega_{2} \in [0.4,0.6]$ and the $\xi \in [3,7]$.
\textbf{Detected noisy correspondences.}
Fig. \ref{fig:case} visualizes some detected mismatched pairs on Flickr30K by ReCon. These pairs exhibit high matching probabilities with local correspondences, yet are correctly identified as mismatched pairs due to their inconsistencies of intra-modal relations.

\section{Conclusion}
This paper introduces a general \textbf{Re}lation \textbf{Con}sistency learning framework, namely \textbf{ReCon}, to effectively mitigate the adverse impact caused by NCs. The main motivation of our ReCon is to \textit{enhance the discriminability of models for true correspondences in noisy multimodal dataset} and thus effectively avoids the wrong supervisions of false correspondences, especially in the presence of hard NCs. Specifically, we leverage the dual constrains, which simultaneously consider the cross- and intra-modal relations, to jointly divide the corrupted training data into different partitions. Extensive experiments conducted on three widely-used cross-modal benchmarks validate the effectiveness and robustness of ReCon in handling both simulated and real-world NCs.

{
    \small
    \bibliographystyle{ieeenat_fullname}
    \bibliography{main}
}


\end{document}